 \documentclass[sigconf]{acmart}
\AtBeginDocument{%
  }

\setcopyright{acmlicensed}
\copyrightyear{2026}
\acmYear{2026}
\setcopyright{cc}
\setcctype{by}
\acmConference[KDD '26]{Proceedings of the 32nd ACM SIGKDD Conference on Knowledge Discovery and Data Mining V.1}{August 09--13, 2026}{Jeju Island, Republic of Korea}
\acmBooktitle{Proceedings of the 32nd ACM SIGKDD Conference on Knowledge Discovery and Data Mining V.1 (KDD '26), August 09--13, 2026, Jeju Island, Republic of Korea}
\acmPrice{}
\acmDOI{10.1145/3770854.3780246}
\acmISBN{979-8-4007-2258-5/2026/08}

\usepackage{amsmath} 
\usepackage{amsthm}
\usepackage{enumitem}
\usepackage{multirow}
\usepackage{booktabs}
\usepackage{makecell}  
\usepackage{array} 
\usepackage{tabularx}

\begin{document}

\title{Live or Lie: Action-Aware Capsule Multiple Instance Learning for Risk Assessment in Live Streaming Platforms}


\author{Yiran Qiao}
\authornote{This work was conducted during Yiran's internship at ByteDance China.}
\authornotemark[3]
\authornotemark[4]
\orcid{0009-0006-0632-0066}
\affiliation{%
  \institution{Institute of Computing Technology, Chinese Academy of Sciences}
  \city{Beijing}
  \country{China}
}
\email{yrqiao@gmail.com}

\author{Jing Chen}
\orcid{0000-0003-2672-4587}
\affiliation{%
  \institution{ByteDance China}
  \city{Hangzhou}
  \country{China}
}
\email{yilan.chan@bytedance.com}

\author{Xiang Ao}
\authornote{Corresponding Author.}
\authornotemark[3]
\authornotemark[4]
\orcid{0000-0001-9633-8361}
\affiliation{%
  \institution{Institute of Computing Technology, Chinese Academy of Sciences}
  \city{Beijing}
  \country{China}
}
\email{aoxiang@ict.ac.cn}

\author{Qiwei Zhong}
\orcid{0000-0002-8517-8072}
\affiliation{%
  \institution{ByteDance China}
  \city{Hangzhou}
  \country{China}
}
\email{huafeng.hf@bytedance.com}

 \author{Yang Liu}
 \authornotemark[3]
\authornotemark[4]
 \orcid{0000-0002-1525-0788}
\affiliation{%
  \institution{Institute of Computing Technology, Chinese Academy of Sciences}
  \city{Beijing}
  \country{China}
}
\email{liuyang2023@ict.ac.cn}

 \author{Qing He}
\authornote{State Key Laboratory of AI Safety, Institute of Computing Technology, Chinese Academy of Sciences.}
\authornote{Also with University of Chinese Academy of Sciences.}
 \orcid{0000-0001-8833-5398}
\affiliation{%
  \institution{Institute of Computing Technology, Chinese Academy of Sciences}
  \city{Beijing}
  \country{China}
}
\email{heqing@ict.ac.cn}

\renewcommand{\shortauthors}{Yiran Qiao et al.}

\begin{abstract}

Live streaming has become a cornerstone of today’s internet, enabling massive real-time social interactions. However, it faces severe risks arising from sparse, coordinated malicious behaviors among multiple participants, which are often concealed within normal activities and challenging to detect timely and accurately.
In this work, we provide a pioneering study on risk assessment in live streaming rooms, characterized by weak supervision where only room-level labels are available. We formulate the task as a Multiple Instance Learning (MIL) problem, treating each room as a bag and defining structured \emph{user–timeslot capsules} as instances. These capsules represent subsequences of user actions within specific time windows, encapsulating localized behavioral patterns.

Based on this formulation, we propose \textbf{AC-MIL}, an Action-aware Capsule MIL framework that models both individual behaviors and group-level coordination patterns. AC-MIL captures multi-granular semantics and behavioral cues through a serial and parallel architecture that jointly encodes temporal dynamics and cross-user dependencies. 
These signals are integrated for robust room-level risk prediction, while also offering interpretable evidence at the behavior segment level.
Extensive experiments on large-scale industrial datasets from Douyin demonstrate that AC-MIL significantly outperforms MIL and sequential baselines, establishing new state-of-the-art performance in room-level risk assessment for live streaming. Moreover, AC-MIL provides capsule-level interpretability, enabling identification of risky behavior segments as actionable evidence for intervention.
The project page is available at: \url{https://qiaoyran.github.io/AC-MIL/}.

\end{abstract}



\maketitle
\newcommand\kddavailabilityurl{https://doi.org/10.1145/3770854.3780246}
\ifdefempty{\kddavailabilityurl}{}{
\begingroup\small\noindent\raggedright\textbf{Resource Availability:}\\
The source code of this paper has been made publicly available at \url{https://doi.org/10.5281/zenodo.18074892}.
\endgroup
}

\section{Introduction}\label{sec:intro}




Live streaming has grown rapidly in recent years and established itself as an essential component of digital services globally. Leveraging real-time interaction and strong social presence, it supports communication, entertainment, and e-commerce on major platforms, including TikTok, Twitch, YouTube, etc.

\begin{figure}[t]
  \centering
\includegraphics[width=0.48\textwidth]{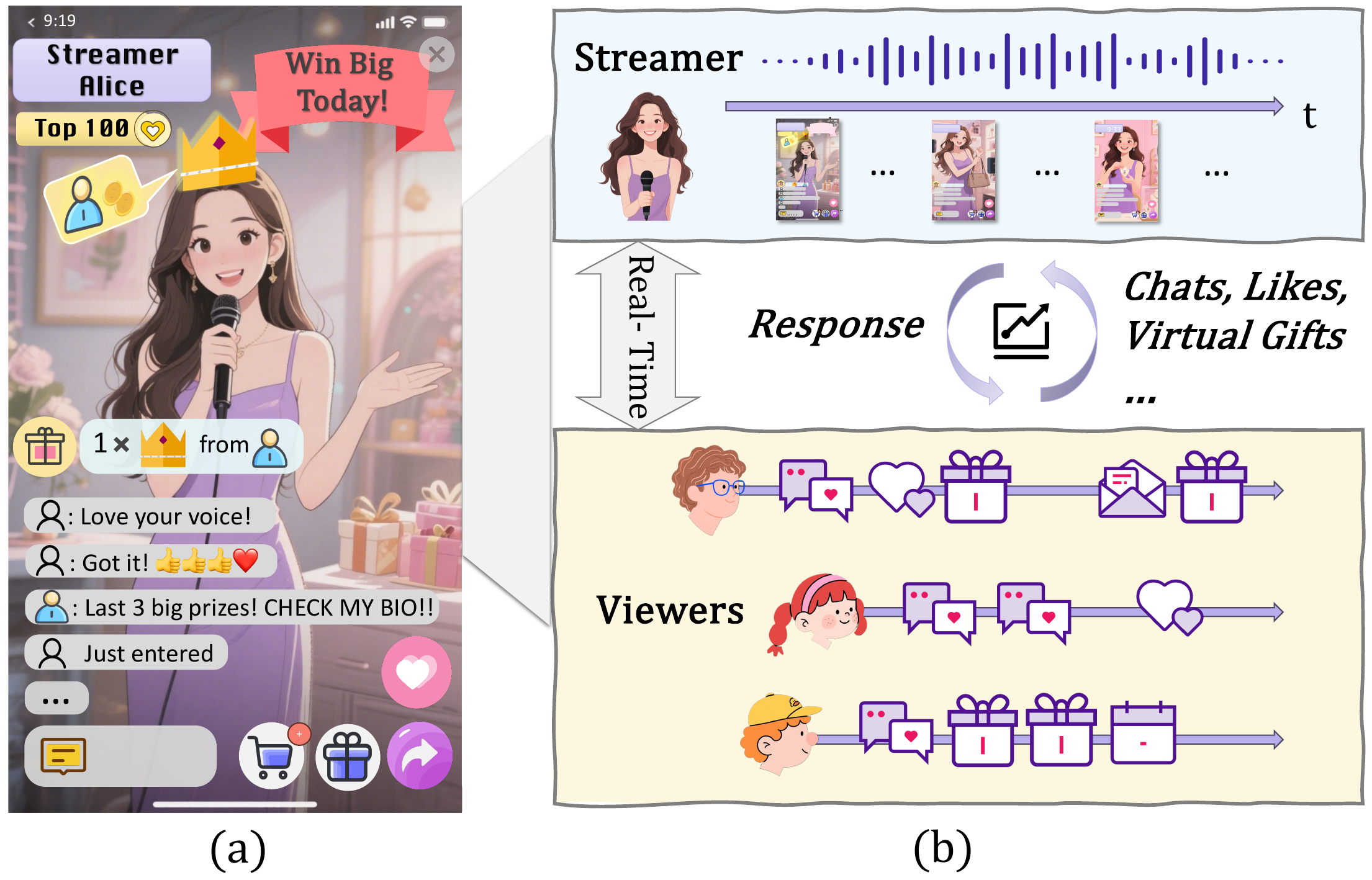}
  \caption{Illustration of a toy live streaming room. (a)~A virtual screenshot showing the streamer continuously delivering verbal and visual content, alongside multiple interactive feature icons for viewers.
(b)~Viewers engage with the streamer in real time through chats, likes, virtual gifts, etc., while the streamer responds viewer interactions.}
    \vspace{-1em}
  \label{fig:intro}
\end{figure}

A live streaming room is the basic unit of such services, where user interactions are centered~(c.f.~Figure~\ref{fig:intro}). Each room involves a streamer who delivers verbal and graphic content and engages with viewers in real time. Viewers participate through chats, virtual gifts, and other interactive actions, creating a dynamic and socially rich environment. 

However, the very features that make live streaming engaging, i.e., its immediacy, interactivity, and open access, also introduce various risks within streaming rooms. Among these, live-streaming fraud is particularly common and harmful. Fraudsters take advantage of the fast-paced, real-time nature of lives to quickly lure viewers into scams.
In many cases, streamers collaborate with planted viewers, who use interactive features to coordinate with the streamer, building false trust and pushing other viewers into quick decisions. Fraudsters commonly adopt scripted routines within live streaming rooms to systematically guide viewers toward external platforms where scams are executed.

Detecting such risks is far from trivial, as the live streaming context poses unique challenges for risk assessment:

(i) \textbf{Complex and indirect signals.}
Live streaming generates large-scale, dynamic user actions across multiple modalities, such as video, audio, text, and metadata.
Moreover, because actual scams typically occur off-platform, the live room may contain only indirect, scripted routines and coordinated interactions as evidence. These trails are hidden within complex behavior data and thus are difficult to detect.

(ii) \textbf{Strict performance requirements.} Risk detection systems must achieve a high recall ratio while maintaining a low false alarm rate to protect users' experience and avoid blocking legitimate activities. In addition, decisions must be explainable and able to pinpoint specific suspicious segments of actions to justify suspensions.

(iii) \textbf{Timeliness constraints.} Risk assessment must function in near real-time to disrupt fraudulent activities before more viewers can be deceived, considering the limited duration of live sessions and the short viewer stay~(e.g., around 30 minutes on average). Moreover, relevant behavioral data begins to accumulate only once a live streaming room starts.

To tackle these challenges, we formulate risk assessment in live streaming platforms as a \textbf{Multiple Instance Learning~(MIL)} problem, where labels are only available for groups of instances, called bags, rather than for individual instances~\cite{dietterich1997solving}.
This formulation is well-suited to bridge the gap between coarse, room-level supervision and the need for fine-grained, interpretable detection of correlated and evolving suspicious behaviors.

Defining instances for MIL in this context is challenging.
Individual user actions are often sparse, diverse, and appear harmless in isolation. However, fraudulent patterns emerge through temporal dependencies and coordinated behaviors across users.
To preserve such contextual cues, we define each instance as a \textbf{\emph{user–timeslot capsule}}, i.e., a subsequence of actions performed by a user within a specific temporal window.
This structured unit captures both user-level behavior and localized temporal dynamics, which are essential for uncovering covert and coordinated risks.

However, this capsule-based MIL setting introduces unique modeling requirements.
Conventional MIL typically assumes independent instances, aggregated through simple pooling~\cite{wu2015deep, wang2018revisiting}.
In contrast, fraudulent behaviors in live streaming often arise from coordinated, temporally aligned interactions across users, which violates the independence assumption.
While some recent works~\cite{ilse2018attention, chen2024timemil, early2024inherently} have explored inter-instance relations or temporal dependencies beyond this assumption, they primarily target uniformly structured, sequential data.
By comparison, our setting features \textbf{\emph{a two-dimensional $\mathbf{U \times T}$ capsule space}}, where each user is a discrete entity and may engage in irregular, collusive behaviors over time.
These characteristics demand a new MIL architecture that can jointly model temporal dynamics, cross-user dependencies, and the irregularity inherent in live streaming interactions.

To this end, we propose an \textbf{\underline{A}}ction-Aware \textbf{\underline{C}}apsule \textbf{\underline{M}}ultiple \textbf{\underline{I}}nstance \textbf{\underline{L}}earning~(\textbf{AC-MIL}) framework for room-level risk detection in live streaming. AC-MIL adopts a hierarchical architecture that combines \textit{serial} and \textit{parallel} modeling to progressively refine room-level behavioral representations.

Specifically, AC-MIL first encodes raw user actions with a context-aware encoder, then constructs localized user–timeslot capsules to capture temporal-spatial interaction patterns. On top of these capsules, a relational capsule reasoning mechanism first models latent correlations among interaction patterns, followed by a dual-view aggregation module that captures complementary user-centric and time-centric perspectives. These modules yield four hierarchical room-level representations fused for accurate risk assessment. 

We evaluate AC-MIL on large-scale industrial datasets from Douyin, a major live streaming platform in China, demonstrating state-of-the-art performance. Additionally, AC-MIL generates capsule-level risk attributions that highlight suspicious user–time patterns, providing interpretable evidence to support actionable risk control.

Our main contributions are summarized as follows:
\begin{itemize}[leftmargin=*,topsep=5pt]

\item We present, to the best of our knowledge, the first pioneering study of risk assessment in live streaming rooms, and formulate it as a Multiple Instance Learning~(MIL) task where each instance is defined as a \emph{user–timeslot capsule}, i.e., a subsequence of user actions within a specific time window.

\item We propose the \textbf{Action-Aware Capsule MIL} (AC-MIL) method,  which captures room-level risk signals across multiple granularities and highlights suspicious user-time capsules for interpretable risk tracing.

\item Extensive experiments on large-scale industrial datasets from Douyin demonstrate that AC-MIL achieves SOTA performance in room-level risk assessment and provides actionable insights for real-world moderation.


\end{itemize}

\section{Related Work}\label{sec:related}
\noindent
 \textbf{Multiple Instance Learning.}
Multiple instance learning~(MIL) is a weakly supervised paradigm where only a single label is provided for a group of instances, known as a bag, while the labels of individual instances remain unknown~\cite{dietterich1997solving}.
Standard MIL methods rely on two assumptions: (i) instances within a bag are independent and permutation-invariant, and (ii) a bag is labeled positive if at least one of its instances is positive. Based on these assumptions, bag-level predictions are typically derived by directly aggregating instance-level predictions, such as through max or mean pooling~\cite{wang2018revisiting,wu2015deep}.

However, these assumptions neglect dependencies among instances~\cite{ilse2018attention, chen2024timemil} and limit the flexibility in modeling the relationship between instance-level signals and bag-level outcomes~\cite{early2024inherently}.
To address these limitations, recent work has relaxed these two assumptions by modeling inter-instance correlations~\cite{chen2024timemil} and introducing soft aggregation mechanisms that allow all instances to adaptively contribute to the bag-level prediction~\cite{ilse2018attention, javed2022additive, early2024inherently}. Such approaches have proven effective for structured data such as time series~\cite{early2024inherently, chen2024timemil, jang2025tail}.

For example, MIL-LET~\cite{early2024inherently} and TimeMIL~\cite{chen2024timemil} apply MIL to time series by capturing temporal patterns and modeling dependencies with order-preserving or time-aware pooling strategies. TAIL-MIL~\cite{jang2025tail} further extends this by supporting two-dimensional instances encoding both temporal positions and variable-level signals in multivariate time series.

Despite sharing the MIL perspective, our work differs substantially from prior time-series-oriented MIL methods. Specifically, (i) unlike regularly sampled time series, our user-timeslot patches are highly uneven due to dynamic user participation; (ii) our minimal units are semantically meaningful user actions rather than numeric values; and (iii) the user dimension carries entity-level significance, as users are potential risk actors, unlike variable channels in multivariate time series. These characteristics yield a two-dimensional instance space ($U \times T$) that requires MIL to capture interactions across both time and user dimensions.\\

\noindent
\textbf{Risk Assessment in Online Platforms.}
Although risk assessment at the level of live streaming rooms has not been explored, other risk management tasks in online platforms have received significant attention. In particular, we review two related areas: \textbf{\textit{automated content moderation}}, which detects toxic user-generated content, and \textbf{\textit{online fraud detection}}, which focuses on identifying malicious transactions or fraudulent users.

\textbf{\textit{Automated content moderation}} focuses on automatively detecting harmful user content~\cite{zannettou2020measuring,kwok2013locate,nguyen2023detecting,markov2023holistic}, with solutions such as Google’s Perspective API for toxic comments~\cite{lees2022new} and Kuaishou’s KuaiMod benchmark for policy-violating short videos~\cite{lu2025vlm}. Differently, our work targets room-level risk assessment based on behavioral patterns and user interactions beyond content alone.

\textbf{\textit{Online fraud detection}} approaches mainly fall into two types: graph-based methods~\cite{dou2020enhancing,huang2022auc,liu2021pick,shi2022h2,li2021live,cheng2025graph}, which model relations among users or transactions to detect fraud; and sequence-based methods, which apply recurrent neural networks~\cite{branco2020interleaved,liu2020fraud} or attention mechanisms~\cite{liu2021intention,guo2018learning} to learn temporal dynamics in user activity. Building on sequence models, hierarchical architectures~\cite{zhu2020modeling,liu2021intention,qiao2024Financial}, pre-training frameworks~\cite{liu2022user,wang2023sequence}, representation enrichment techniques~\cite{qiao2025online,xiao2024vecaug} are adopted to extract user intentions.

Specifically, Taobao~\cite{li2021live} detects fraudulent transactions in e-commerce live streaming rooms, essentially following the transaction-level fraud detection paradigm only using behavioral data from the live scene.
In contrast, we target comprehensive risk assessment at the live streaming room level, leveraging multimodal information of user actions and capturing richer temporal and interaction dynamics beyond single events.

\section{Problem Formulation}\label{sec:concept}
\subsection{Business Setting}
Our goal is to enable early detection of risky live streaming rooms that may violate platform policies or regulations, while providing fine-grained, localized signals as evidence for enforcement.
Unlike pure content moderation, we focus on user behaviors and interactions, which are key to uncovering coordinated or collusive activities that content analysis may miss. We formulate this as a binary classification task to identify risky rooms and safeguard the platform ecosystem.
\subsection{Live Streaming Room Definition}

To model live streaming risk assessment as an MIL problem, we first describe the structural composition of a live streaming room at the finest granularity, namely, the action level.

\begin{definition}
\textbf{(Role-Based Live Streaming Action)}  
An action occurring in a live streaming room $r \in \mathcal{R}$ is defined as a 4-tuple $\alpha = (u, t, a, x)$, where:
\begin{itemize}[leftmargin=*,topsep=5pt]
    \item $u \in \mathcal{U}$ is the user ID, where $\mathcal{U}$ denotes the user set.
    \item $t \in \mathcal{T}$ is the action timestamp.
    \item $a \in \mathcal{A}$ is the action ID, where $\mathcal{A} = \mathcal{A}^{\texttt{streamer}} \cup \mathcal{A}^{\texttt{viewer}}$ represents the action type set.
    \item $x \in \mathbb{R}^d$ is the textual information associated with the action, e.g., the content of a comment; for actions without text, this is set to \texttt{none}. 
\end{itemize}
\end{definition}
\begin{definition}
\label{def:room seq}
\textbf{(Live Streaming Room)}  
A live streaming room $r$ within time window $[0, T]$ is defined as:
\begin{equation}
    R_r^{[0,T]} = \Big( \mathcal{U}_{r},\; S_r^{[0,T]} \Big),
\end{equation}
where $\mathcal{U}_{r} = \{ u_r^{\texttt{s
}} \} \cup U_r^{\texttt{v}, [0,T]}$ denotes the user set in room $r$, consisting of the unique streamer user $u_r^{\texttt{s}}$ and the viewer users $U_r^{\texttt{v}, [0,T]}$. And $S_r^{[0,T]} = \big\{ \alpha_i \;|\; \alpha_i = (u_i, t_i, a_i, x_i),\; u_i \in \mathcal{U}_{r},\; 0 \le t_i -t_0\le T \big\}$ is the sequence of all action events in room $r$ during $[0, T]$ after the room starts at $t_0$, where $N^{\mathrm{a}}_r\triangleq |S_r^{[0,T]}| $ denotes the total number of actions. 
For brevity, in the following sections, we omit the room index $r$ in symbols related to entities such as users and action sequences, while keeping it in the overall room representation.
\end{definition}

\subsection{$U \times T$ Capsule-Level MIL}

To facilitate the identification of risk-sensitive patterns localized in specific users or timeslots, we partition the live streaming room into \emph{user-timeslot capsules}. Let the time window $[0, T]$ be divided into $K$ consecutive timeslots $\{\mathcal{T}_k\}_{k=1}^K$. For each user $u \in \mathcal{U}$ and timeslot $\mathcal{T}_k$, we define a capsule as:
\begin{equation}
    C_{u,k} = \big[\, \alpha_i \;|\; \alpha_i = (u, t_i, a_i, x_i),\; t_i \in \mathcal{T}_k \big].
\end{equation}

Each capsule $C_{u,k}$ is defined as a time-ordered sub-sequence of actions performed by user $u$ within timeslot $\mathcal{T}_k$, which is an instance. Thus, a live streaming room $r$ is represented as a bag of capsules:
\begin{equation}
    \mathcal{B}_r = \left\{\, C_{u,k} \;\middle|\; u \in \mathcal{U},\; k = 1,\ldots,K \right\}.
\end{equation}

We formulate live streaming risk detection as an MIL problem, where each bag $\mathcal{B}_r$ is associated with a binary label $y_r \in \{0,1\}$ indicating whether the room is risky~($y_r=1$).

\section{Methodology}\label{sec:method}
\subsection{Overview of AC-MIL}
\begin{figure*}[htbp]
  \centering
  \includegraphics[width=\textwidth]{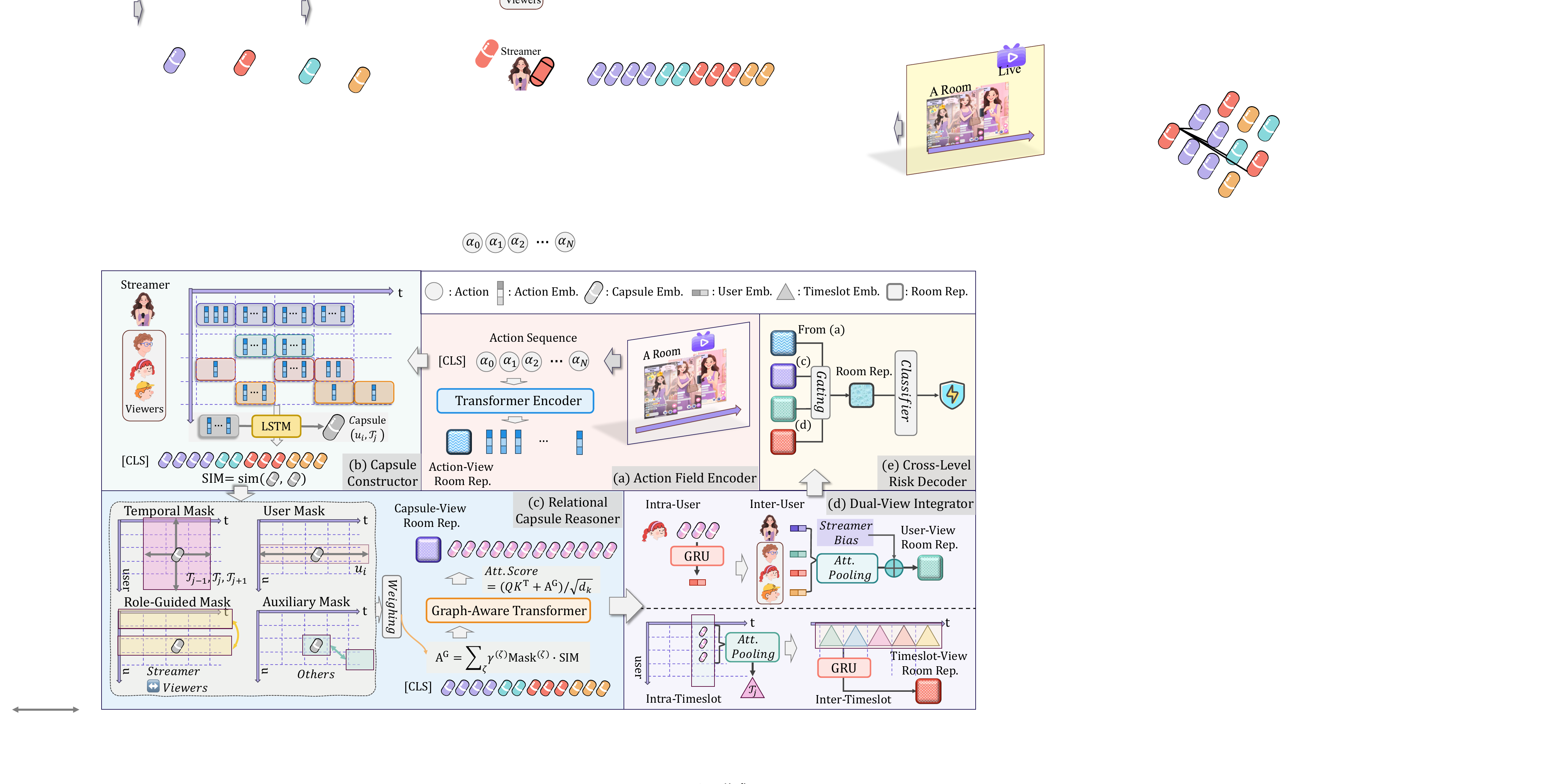}

  \caption{Overview of AC-MIL, a hierarchical serial-parallel framework that models raw user actions to produce room-level risk predictions.
Key components include: (a) Action Field Encoder for contextualizing raw behaviors; 
(b) Capsule Constructor that structures actions into interpretable user–time capsules; 
(c) Relational Capsule Reasoner modeling dependencies via graph-aware self-attention~(also providing capsule-level interpretability); 
(d) Dual-View Integrator capturing user- and time-centric views; 
and (e) Cross-Level Risk Decoder aggregating multi-granular contextual signals for final risk scoring. 
}
  \label{fig:overview}
\end{figure*}

As depicted in Figure~\ref{fig:overview}, AC-MIL is built upon a hierarchical serial-parallel architecture that progressively refines behavioral representations from atomic user actions to room-level risk predictions, while integrating multi-view contextual signals in parallel.

First, \textbf{(a)~Action Field Encoder} transforms the raw multimodal user behavior stream into globally contextualized representations. These embeddings are then organized through \textbf{(b)~Capsule Constructor}, which segments the action space into structured capsules, i.e., localized behavioral units formed over user–time grids. These capsules serve as interpretable intermediate representations that capture rich local semantics.

Then, \textbf{(c)~Relational Capsule Reasoner} employs a graph-aware self-attention mechanism to discover latent structures in user interactions and behavioral patterns, modeling intricate dependencies and collaborative anomalies across capsules, and also produces capsule-level risk attributions for localized interpretability.
Subsequently, \textbf{(d)~Dual-View Integrator} encodes user-centric and time-centric contextual views in parallel, capturing both structural roles and temporal rhythms to provide complementary perspectives that enhance expressiveness.
Finally, \textbf{(e)~Cross-Level Risk Decoder} aggregates information from the action, capsule, user, and timeslot levels, producing final room-level risk predictions.
This action-aware capsule MIL approach effectively integrates multi-granular contextual signals with localized interpretability to enable reliable risk assessment.
\subsection{Capsule Structuring}
We begin by introducing the initial stage of AC-MIL, namely the capsule structuring process, which comprises the Action Field Encoder and the Capsule Constructor. The Action Field Encoder transforms raw user actions into globally contextualized representations. Subsequently, the Capsule Constructor reorganizes the action embeddings into user–time grids to form capsules, laying the foundation for subsequent capsule-based reasoning.

\noindent
\textbf{Action Field Encoder.}
Given the room-level action sequence $S^{[0,T]} = \{ \alpha_i = (u_i, t_i, a_i, x_i), i \in [1,N^{\mathrm{a}}] \}$, 
we first transform each action into a vector representation. 
Specifically, for each action $\alpha_i$, we compute
 $
    \mathbf{e}_i = \big[\, \mathbf{e}_{a_i} \;\|\; \mathrm{Proj}(x_i)\, \big]$,
where $\mathbf{e}_{a_i}$ is the learnable embedding of the action ID $a_i$, and $\text{Proj}(\cdot)$ is a learnable projection layer that maps the action’s pre-encoded textual feature $x_i \in \mathbb{R}^d$ into a fixed-size vector with dimension $d_k$.
To capture global action-level dependencies, we flatten all action embeddings into a single sequence, prepend a learnable $\texttt{[CLS]}$ token, and feed it into a Transformer encoder~\cite{vaswani2017attention}:
\begin{equation}
    \mathbf{H}^{\mathrm{a}} = \mathrm{TransformerEncoder}\big(\texttt{[CLS]},\; \mathbf{e}_1, \mathbf{e}_2, \dots, \mathbf{e}_{N^{\mathrm{a}}}\big).
\end{equation}
The output embedding of $\texttt{[CLS]}$, denoted as $\mathbf{h}^{\mathrm{a}}_r = \mathbf{H}^\mathrm{a}[0]$, 
serves as the \emph{action-level room representation}, capturing global contextual semantics across the entire stream.
The remaining token embeddings~(still denoted as  $\{\mathbf{e}_i, i \in [1,N^\mathrm{a}]\}$ for brevity) preserve fine-grained behavioral patterns 
and are reorganized into structured capsules over user–time grids for higher-level reasoning.

\noindent
\textbf{Capsule Constructor.}  
Risk in live streaming rooms is often localized to specific users and brief time intervals. Such fine-grained \emph{user–timeslot} patterns provide critical cues for detecting anomalous or coordinated behaviors that global aggregation may neglect. To capture these localized signals, we partition the room into \emph{user–timeslot capsules}.

Recall that each capsule $C_{u,k}$ contains a short time-ordered sequence of a single user's actions within a specific timeslot: 
$
    C_{u,k} = \big[\, \alpha_i \;|\; \alpha_i = (u, t_i, a_i, x_i),\; t_i \in \mathcal{T}_k \big].
$
We employ a Long Short-Term Memory~(LSTM) model~\cite{graves2012long} to learn this sequence and use the final hidden state as the capsule embedding:
\begin{equation}
\mathbf{c}_{u,k} = \mathrm{LSTM}\left( \left[\mathbf{e}_i\right]_{\alpha_i \in C_{u,k}} \right)[-1],
\end{equation}
where $\mathbf{e}_i$ is the embedding of action $\alpha_i$, obtained from the Action Field Encoder.
By decomposing the room into \emph{user–timeslot capsules}, we form localized units for modeling cross-capsule interactions and detecting collaborative anomalies.

\subsection{Relational Capsule Reasoner}
While \emph{user–timeslot capsules} capture localized behavioral patterns, many risk signals in live streaming rooms emerge from complex interactions across users and timeslots. For example, coordinated fraud often involves multiple users acting in different time windows yet exhibiting correlated behaviors. 

\noindent
\textbf{Adaptive Graph Construction.}  
To uncover such collaborative anomalies, we first build an adaptive global graph over all capsules.
Specifically, we unify all \emph{user–timeslot capsules} into a global node set $\mathcal{V} = \{v_i\}_{i=1}^{N^\mathrm{c}}$, where each node $v_i$ corresponds to the capsule embedding $\mathbf{c}_{u_i,k_i}$ of user $u_i$ in timeslot $\mathcal{T}_{k_i}$, with $N^\mathrm{c} \triangleq|\mathcal{B}_r|$ denoting the total number of capsules within the room.

To quantify the semantic affinity between capsules, we compute a similarity matrix as: $
\mathbf{SIM}_{ij} = \mathrm{GELU}\bigl(\mathbf{c}_i^\top \mathbf{c}_j\bigr),
$
where $\mathbf{c}_i$ and $\mathbf{c}_j$ are the capsule embeddings corresponding to nodes $v_i$ and $v_j$, respectively. This similarity matrix serves as a foundation for modeling pairwise relations in the graph.


However, semantic similarity alone may fail to capture critical structural dependencies in live streaming, such as temporal continuity in user behavior or streamer–audience interactions. To encode such diverse relational semantics, we define a set of relation types $\mathcal{R} = \{\mathrm{t}, \mathrm{u}, \mathrm{r}, \mathrm{a}\}$, each represented by a binary mask matrix $\mathbf{M}^{(\zeta)} \in \{0,1\}^{N^\mathrm{c} \times N^\mathrm{c}}, \zeta \in \mathcal{R}$. These masks capture distinct patterns of interaction, which we detail as:

\begin{itemize}[leftmargin=*,topsep=5pt]
\item \textbf{Temporal relations ($\mathrm{t}$)} connect capsules from the same or neighboring timeslots, modeling possible co-occurring or sequentially related interactions within a short time window.
\item \textbf{User relations ($\mathrm{u}$)} connect all capsules that belong to the same user across different timeslots, capturing user-specific behavioral patterns or evolving tendencies.
\item \textbf{Role-guided relations ($\mathrm{r}$)} link the streamer’s capsules with those of all viewers, modeling role-induced interaction patterns that characterize anchor–audience dynamics.
\item \textbf{Auxiliary relations ($\mathrm{a}$)} connect capsule pairs not covered by any of the above relations, serving as a residual channel to detect potentially overlooked collaborative anomalies.
\end{itemize}
Formally, the relations are encoded by the following mask matrices, where $k_i$ and $k_j$ denote the timeslots of capsules $i$ and $j$, $u_i$ and $u_j$ denote their corresponding users, and $u^{\texttt{s}}$ represents the streamer:
\begin{equation}
\begin{aligned}
\mathbf{M}^{\mathrm{t}}_{ij} &= \mathbf{1}\bigl[|k_i - k_j| \le 1\bigr], \\
\mathbf{M}^{\mathrm{u}}_{ij} &= \mathbf{1}\bigl[u_i = u_j\bigr], \\
\mathbf{M}^{\mathrm{r}}_{ij}
&= \mathbf{1}\bigl[
\bigl(u_i = u^{\texttt{s}} \wedge u_j \neq u^{\texttt{s}}\bigr)\\
    &\quad\;\;\;\lor 
\bigl(u_j = u^{\texttt{s}} \wedge u_i \neq u^{\texttt{s}}\bigr)
\bigr],\\
\mathbf{M}^{\mathrm{a}}_{ij} &= 1 - \min\bigl(1, \mathbf{M}^{\mathrm{t}}_{ij} + \mathbf{M}^{\mathrm{u}}_{ij} + \mathbf{M}^{\mathrm{r}}_{ij} \bigr).
\end{aligned}
\label{eq:masks}
\end{equation}

We first compute the unnormalized adaptive relation-aware adjacency matrix $\mathbf{\tilde{A}}^{\mathrm{G}}$ as a weighted sum over relation types, with $\gamma^{(\zeta)}$ as learnable parameters controlling the importance of each relation:
\begin{equation}
\mathbf{\tilde{A}}^{\mathrm{G}}_{ij} = \sum_{\zeta \in \mathcal{R}} \gamma^{(\zeta)}  \cdot \mathbf{M}^{(\zeta)}_{ij} \cdot \mathbf{SIM}_{ij}, \quad
\mathbf{A}^G_{ij} = \frac{\exp\left(\mathbf{\tilde{A}}^{\mathrm{G}}_{ij}\right)}{\sum_{j'=1}^N \exp\left(\mathbf{\tilde{A}}^{\mathrm{G}}_{i j'}\right)},
\end{equation}
where $\mathbf{A}^{\mathrm{G}}$ is the row-wise normalized adjacency matrix used for downstream capsule relational reasoning.


Note that these relation masks are not mutually exclusive; an edge can simultaneously belong to multiple relation types and thus receive accumulated weights. For example, capsules involving the streamer and a user in adjacent timeslots may be emphasized through both the temporal and role-guided relations.

\noindent
\textbf{Cross-Capsule Relational Learning.}
With the normalized relation-aware graph $\mathbf{A}^{\mathrm{G}}$ constructed, we next inject this structural prior into capsule representation learning. To this end, inspired by Graph Transformer works~\cite{shehzad2024graph}, we tailor a Transformer-based architecture with a graph-aware self-attention mechanism, which integrates both semantic similarities and topological priors.

We begin with the capsule embedding sequence, obtained by flattening the user–timeslot grid first by user, then by timeslot: $\mathbf{C} = [\mathbf{c}_1,\mathbf{c}_2,\dots, \mathbf{c}_{N^\mathrm{c}} \,] \in \mathbb{R}^{N^\mathrm{c}\times d_k}$. A learnable [$\texttt{CLS}$] token is prepended to form the input sequence as $\widetilde{\mathbf{C}} = \bigl[\, [\texttt{CLS}] \,,\; \mathbf{C} \bigr]$.

Accordingly, the adjacency matrix $\mathbf{A}^{\mathrm{G}}$ is expanded to include connections between $[\texttt{CLS}]$ and all capsules, and we control the connection strength by a hyperparameter $\gamma^{[\texttt{CLS}]}$. Then, the standard multi-head projections are computed as:
$\mathbf{Q} = \widetilde{\mathbf{C}} \mathbf{W}^Q,
\mathbf{K} = \widetilde{\mathbf{C}} \mathbf{W}^K,
\mathbf{V} = \widetilde{\mathbf{C}} \mathbf{W}^V.
$
Subsequently, the graph-aware self-attention integrates both feature similarity and topological priors as:
\begin{equation}
\label{eq:att}
\mathrm{Attention}(\mathbf{Q}, \mathbf{K}, \mathbf{V}, \mathbf{A}^{\mathrm{G}}\}
=
\mathrm{Softmax}\left(
\frac{\mathbf{Q}\mathbf{K}^\top + \mathbf{A}^{\mathrm{G}}}{\sqrt{d_k}}
\right)\mathbf{V}.
\end{equation}

This formulation enables capsules to interact based on both semantic similarity and diverse topological priors, i.e., temporal, user, role-guided, and residual relations, so that even capsules outside predefined relations, yet with strong semantic affinity, can still be emphasized. Meanwhile, the $[\texttt{CLS}]$ token aggregates global contextual signals for room-level reasoning.

Consequently, we have
$
\mathbf{H}^\mathrm{c}
=
\mathrm{GraphAwareTransformer}\bigl(\widetilde{\mathbf{C}}, \mathbf{A}^{\mathrm{G}}\bigr),
$ with the tailored self-attention mechanism. 
$\mathbf{H}^\mathrm{c}$ contains refined capsule embeddings, and $\mathbf{h}^\mathrm{c}_r = \mathbf{H}^\mathrm{c}[0]$ serves as a global \emph{capsule-level room representation} for downstream cross-level risk prediction.

\subsection{Dual-View Integrator}
Live streaming risks involve user-specific behavioral trajectories and temporal accumulation of suspicious signals. We restructure the refined capsule embeddings into user- and timeslot-aligned views to enable complementary modeling of inter-user relations and time-evolving risk patterns in parallel.

\noindent
\textbf{User-View Modeling.} 
For \emph{intra-user modeling}, we collect all capsule embeddings of each user across timeslots and encode their sequential dynamics via a Gated Recurrent Unit~(GRU)~\cite{cho2014properties}. Formally, for user $u$, the input sequence $\{\mathbf{c}_{u,k}\}_{k=1}^{K_u}$ is fed into a single-user GRU: 
\begin{equation}
\mathbf{h}_{u,k} = \mathrm{GRU}\bigl(\mathbf{c}_{u,k}, \; \mathbf{h}_{u,k-1}\bigr),
\end{equation}
where $\mathbf{h}_{u,k}$ denotes the hidden state at timeslot $k$. The last hidden state $\mathbf{h}_{u,K_u}$ is taken as the \emph{user embedding} $\mathbf{h}^{u}$. 

For \emph{inter-user modeling}, user embeddings within the room are aggregated via attention pooling that assigns each user $u$ a score:
\begin{equation}
\label{eq:user_attention}
\begin{gathered}
\alpha_{u} 
= \mathrm{Softmax}\Bigl(
    \mathbf{w}^\top 
    \bigl[
        \tanh\bigl(\mathbf{W}_v\, \mathbf{h}^{u}\bigr)
        \odot 
        \sigma\bigl(\mathbf{W}_u\, \mathbf{h}^{u}\bigr)
    \bigr]
    + \mathbf{1}[u = u^\texttt{s}] \cdot b^{\texttt{s}}
\Bigr), \\
\mathbf{h}^u_r = \sum_{u=1}^{|\mathcal{U}|} \alpha_u \, \mathbf{h}^{u},
\end{gathered}
\end{equation}
where $\odot$ denotes element-wise multiplication, $\sigma(\cdot)$ is the sigmoid activation, and $\mathbf{w}$, $\mathbf{W}_v$, $\mathbf{W}_u$, and $b^{\texttt{s}}$ are learnable parameters. The bias $b^{\texttt{s}}$ is added exclusively to the streamer $u^{\texttt{s}}$'s attention score, emphasizing their central role in room interactions.
This cross-user aggregation highlights key participants with consistently abnormal behaviors or coordinated patterns such as collusion, thus enriching the \emph{user-level room representation} $\mathbf{h}^u_r$.

\noindent
\textbf{Timeslot-View Modeling.}
For \emph{intra-timeslot modeling}, let $\{\mathbf{c}_{i}\}_{i \in \mathcal{T}_k}$ denote the set of capsule embeddings within timeslot $\mathcal{T}_k$. We apply an attention pooling mechanism analogous to Eq.~(\ref{eq:user_attention}), but without the streamer bias term, to compute the timeslot embedding:
\begin{equation}
\alpha_{i}' 
= \mathrm{Softmax}\Bigl(
    \mathbf{w'}^\top 
    \bigl[
        \tanh\bigl(\mathbf{W'}_V\, \mathbf{c}_{i}\bigr)
        \odot 
        \sigma\bigl(\mathbf{W'}_U\, \mathbf{c}_{i}\bigr)
    \bigr]
\Bigr), \quad
\mathbf{t}_{k} = \sum_{i \in \mathcal{T}_k} \alpha_i' \, \mathbf{c}_{i},
\end{equation}
where $\mathbf{w'}, \mathbf{W'}_V, \mathbf{W'}_U$ are learnable parameters, and $\mathbf{t}_k$ is the resulting timeslot embedding summarizing key user interactions in that interval.

For \emph{inter-timeslot modeling}, the sequence of timeslot embeddings $\{\mathbf{t}_k\}_{k=1}^T$ is fed into a GRU to capture the temporal progression of risk signals across the live streaming session. 
The final hidden state of the GRU is taken as the global \emph{timeslot-level room representation} $\mathbf{h}^\mathrm{t}_r$, encoding cumulative dynamics and enabling the detection of sequential or bursty anomalies spanning multiple timeslots.

\subsection{Cross-Level Risk Decoder}
\textbf{Representation Fusion and Classification.} 
After obtaining room-level representations from multiple semantic levels, we fuse these complementary signals via a gating mechanism to produce the final room embedding for classification.
Specifically, we apply separate gates to each level’s representation:
\begin{equation}
\begin{gathered}
g^{\mathrm{a}} = \sigma\bigl( \mathrm{MLP}_{\mathrm{a}}(\mathbf{h}^{\mathrm{a}}_r) \bigr),\quad
g^{\mathrm{c}} = \sigma\bigl( \mathrm{MLP}_{\mathrm{c}}(\mathbf{h}^{\mathrm{c}}_r) \bigr),\\
g^{\mathrm{u}} = \sigma\bigl( \mathrm{MLP}_{\mathrm{u}}(\mathbf{h}^{\mathrm{u}}_r) \bigr),\quad
g^{\mathrm{t}} = \sigma\bigl( \mathrm{MLP}_{\mathrm{t}}(\mathbf{h}^{\mathrm{t}}_r) \bigr),
\end{gathered}
\end{equation}
where each $g$ denotes a scalar weight learned for its corresponding representation, and each 2-layer $\mathrm{MLP}{(\cdot)}$ is independently parameterized to allow flexible balancing of multi-level signals.
The final room-level representation $\mathbf{h}_r$ is computed as a weighted sum:
\begin{equation}
\mathbf{h}_r =
g^{\mathrm{a}} \cdot \mathbf{h}^{\mathrm{a}}_r
\;+\;
g^{\mathrm{c}}  \cdot \mathbf{h}^{\mathrm{c}}_r
\;+\;
g^{\mathrm{u}} \cdot \mathbf{h}^{\mathrm{u}}_r
\;+\;
g^{\mathrm{t}} \cdot \mathbf{h}^{\mathrm{t}}_r.
\end{equation}

Finally, $\mathbf{h}_r$ is fed into a classifier~(implemented as a 2-layer MLP) for room-level risk classification: 
$\hat{\mathbf{y}}_r = \mathrm{Classifier}(\mathbf{h}_r).$
This cross-level decoding enables the model to adaptively integrate signals across multiple granularities, capturing both fine-grained action dynamics and high-level collaborative or temporal risk patterns for comprehensive room-level risk assessment.

\noindent
\textbf{Optimization Objective.} 
We train the AC-MIL model using the binary cross-entropy~(BCE) loss summed over all training samples:
$\mathcal{L} = - \sum_{r=1}^N \bigl[ y_r \log(\hat{y}_r) + (1 - y_r) \log(1 - \hat{y}_r) \bigr],$
where $N$ is the number of training rooms, $y_r \in \{0,1\}$ is the ground-truth label, and $\hat{y}_r \in (0,1)$ is the predicted risk score.

\noindent
\textbf{Localized Interpretability.} 
With reference to Eq.~\ref{eq:att}, we treat the [$\texttt{CLS}$] token as the query and examine its attention weights over all capsules. Specifically, we extract the attention distribution $\boldsymbol{\alpha}^{[\texttt{CLS}]} \in \mathbb{R}^{1 \times N^\mathrm{c}}$ from $\mathbf{A}^{\mathrm{G}}_{\texttt{[CLS]}, :}$ which denotes the first row of the attention matrix:
\begin{equation}
\label{eq:att score}
\boldsymbol{\alpha}^{[\texttt{CLS}]} = \mathrm{Softmax} \left( 
\frac{ \mathbf{Q}_{\texttt{[CLS]}} \mathbf{K}^\top + \mathbf{A}^{\mathrm{G}}_{\texttt{[CLS]}, :} }{ \sqrt{d_k} } 
\right),
\end{equation}
where each entry $\boldsymbol{\alpha}^{[\texttt{CLS}]}_j$ quantifies the contribution of capsule $j$ to the aggregated room-level representation, providing capsule-level interpretability for actionable suspension decisions.

\section{Experiments}\label{sec:exp}

\begin{table*}[t]
\centering
\caption{Performance comparison between AC-MIL and baselines. The best and second-best results of the baselines are bold and underlined, respectively. The `*' indicates that all the performance improvements over the best baselines are statistically significant (p-value $<$ 0.05).}
\label{tab:exp:main}

\resizebox{\textwidth}{!}{
\begin{tabular}{@{}c|c|cccc|cccc@{}}   
\toprule
\multicolumn{2}{@{}c|}{\multirow{3}{*}{\textbf{Methods}}} & \multicolumn{4}{c|}{\textbf{May Dataset}} & \multicolumn{4}{c@{}}{\textbf{June Dataset}} \\ 
\cmidrule(lr){3-6} \cmidrule(lr){7-10}
\multicolumn{2}{@{}c|}{} & \makecell{PR-AUC $\uparrow$} & F1-score $\uparrow$ & R@0.1FPR $\uparrow$ & FPR@0.9R $\downarrow$
& PR-AUC $\uparrow$ & F1-score $\uparrow$ & R@0.1FPR $\uparrow$ & FPR@0.9R $\downarrow$ \\ 
\midrule
\multirow{3}{*}{{\shortstack{\textit{Sequence}\\ \textit{Models}}}}
& Transformer   & 0.7189 & 0.6668 & 0.8394 & 0.1580  & 0.6801 & 0.6341  & 0.8225 & 0.1565 \\
& Reformer     & 0.7203 & 0.6752 & 0.8575 & 0.1436 & 0.6911  & 0.6395 & 0.8104 & 0.1760 \\
& Informer   & 0.7246 & 0.6708 & 0.8438 & 0.1555 & 0.6879 & 0.6391  & 0.8375 & 0.1601 \\
\midrule
\multirow{6}{*}{{\shortstack{\textit{MIL}\\ \textit{Methods}}}}
& mi-NET    & 0.7276 & 0.6769 & 0.8560  & \underline{0.1320} & 0.6911 & 0.6406 & 0.8225 & 0.1673  \\
& AtMIL    & 0.7252 & 0.6763 & 0.8550 & 0.1441 & 0.6952 & 0.6519  & 0.8415 & 0.1479   \\
& AdMIL     & 0.7196 & 0.6781 & 0.8389 & 0.1568 & 0.6837 & 0.6491 & 0.8225 & 0.1565 \\
& MIL-LET   & 0.7241 & 0.6749 & 0.8546 & 0.1418 & 0.6942 & \underline{0.6528} & 0.8455 & 0.1499  \\
& TimeMIL   & \underline{0.7353} & \underline{0.6790} & \underline{0.8599} & 0.1436  & 0.6963 & 0.6471  & \underline{0.8495} & \underline{0.1367} \\
& TAIL-MIL   & 0.7316 & 0.6785 & 0.8570 & 0.1341  & \underline{0.7029} & 0.6509 & 0.8205 & 0.1555 \\
\midrule
\multicolumn{2}{@{}c|}{\textbf{AC-MIL}(ours)}   & \textbf{0.7676*} & \textbf{0.7002*} & \textbf{0.8722*} & \textbf{0.1260*} & \textbf{0.7311*} & \textbf{0.6777*} & \textbf{0.8546*} & \textbf{0.1345*} \\
\multicolumn{2}{@{}c|}{\textbf{Best Improv.}} &
\textit{+4.4\%} & \textit{+3.1\%} & \textit{+1.4\%} & \textit{-4.8\%} & 
\textit{+4.0\%} & \textit{+3.8\%} & \textit{+1.0\%} & \textit{-1.6\%} \\ 
\bottomrule
\end{tabular}
}  
\end{table*}
In this section, we validate the effectiveness of AC-MIL on large-scale real-world datasets from a major live streaming platform, with the aim of addressing the following research questions:
\begin{itemize}[leftmargin=*,topsep=5pt]
    \item \textbf{RQ1}: How does AC-MIL perform in live streaming risk assessment compared to existing methods?
    \item \textbf{RQ2}: How do the different components of AC-MIL contribute to its overall effectiveness?
    \item \textbf{RQ3}: Can AC-MIL effectively detect typical scripted suspicious behaviors in live streaming scenarios?
    \item \textbf{RQ4:} Does AC-MIL learn meaningful room representations that separate risky and normal patterns?
    \item \textbf{RQ5:} Can AC-MIL outperform deployed online models in real-world live streaming environments?
    \item \textbf{RQ6:} How sensitive is AC-MIL to key hyperparameters used in capsule construction?
\end{itemize}

\subsection{Experimental Setup}
\textbf{Datasets.} We conducted experiments on two large-scale industrial datasets from Douyin, a major live streaming platform in China\footnote{All data collection adhered to the platform's security and privacy regulations.}, denoted as \textbf{May} and \textbf{June}. 
Each sample is a raw action sequence from a live streaming room (cf. Definition~\ref{def:room seq}), truncated to the first 30 minutes to support early-stage risk detection, with each timeslot set to 100 seconds. We further filter out inactive users who only entered the room without performing any other actions.
Besides, we retain all positive instances and sample negatives at a 1:10 ratio, resulting in a positive class proportion of approximately 9.09\%. For each room, we select the top 50 most active viewers based on total action count to construct the action sequence. 

Regarding the time spans, the \textbf{May} dataset contains training data from 05/20/2025 to 06/03/2025, validation data from 06/11/2025 to 06/12/2025, and test data from 06/13/2025 to 06/14/2025. The \textbf{June} dataset includes training data from 06/04/2025 to 06/10/2025, validation data on 06/15/2025, and test data on 06/16/2025. The basic statistics are displayed in Table~\ref{table:exp:dataset}. More details are provided in Appendix~\ref{sec:app:data}.
\begin{table}[b]
\caption{The basic statistics of the May and June datasets~(Avg.time in minutes).}
\label{table:exp:dataset}
\centering
\resizebox{\linewidth}{!}{
  \begin{tabular}{c|c|cccc} 
  \toprule
   &  & \#Rooms & \#Avg.actions & \#Avg.users &Avg.time \\ 
  \midrule
  \multirow{3}{*}{\textbf{May}} & train & $176,347$ &709  & 35 & 30.0 \\ 
   & val & $23,562$ & 704 & 36 & 29.6 \\ 
   & test & $22,462$ & 740 & 37 & 29.7 \\ 
  \midrule
  \multirow{3}{*}{\textbf{June}} & train & $79,552$ & 700 & 36 & 30.0 \\ 
   & val & $10,934$ & 767 & 40 & 29.1 \\ 
   & test & $10,967$ &725 & 37 &29.1  \\ 
  \bottomrule
  \end{tabular}
}
\end{table}

\noindent
\textbf{Baselines.}
As there is no existing method tailored for room-level risk assessment in live streaming, we adopt strong baselines from two relevant directions. Given the sequential nature of the data and our use of an MIL framework, we consider (i) \textit{Sequence Models} to directly model the action sequences of rooms: \textbf{Transformer}~\cite{vaswani2017attention}, \textbf{Reformer}~\cite{kitaev2020reformer}, and \textbf{Informer}~\cite{zhou2021informer}.
(ii) \textit{MIL methods} to make room-level predictions via instance aggregation: \textbf{mi-NET}~\cite{wang2018revisiting}, \textbf{Attention MIL~(AtMIL)}~\cite{ilse2018attention}, \textbf{Additive MIL~(AdMIL)}~\cite{javed2022additive}, 
\textbf{MIL-LET}~\cite{early2024inherently},  \textbf{TimeMIL}~\cite{chen2024timemil}, and \textbf{TAIL-MIL}~\cite{jang2025tail}. More details are introduced in Appendix~\ref{sec:app:baselines}.

\noindent
\textbf{Implementation Details.}
Our implementation is based on Python 3.11.2 and PyTorch 2.6.0. The hyperparameter $\gamma^{[\texttt{CLS}]}$ is tuned from $\{1.0,1.5,2.0\}$. To mitigate overfitting, we adopt early stopping with a patience of 20 epochs, and the maximum number of training epochs is set to 100. Except for the Graph-Aware Transformer, which uses a single layer, all other sequence models, including LSTM, GRU, and the Transformer components in both AC-MIL and baseline methods, use two layers. The number of attention heads is set to 8 for all Transformer-based models. The dropout rate is set to 0.1, and all embedding dimensions are fixed at 128. We use a batch size of 128 and train with the AdamW optimizer, using a learning rate of 0.0001 and a weight decay of 0.0001.

\noindent
\textbf{Evaluation Metrics.}
We adopt \textbf{PR-AUC}, \textbf{F1-score}, \textbf{R@0.1FPR}, and \textbf{FPR@0.9R} as evaluation metrics. PR-AUC and F1-score reflect the balance between precision and recall, which is crucial for risk detection under severe class imbalance. While ROC-AUC is widely used, it can be overly optimistic in this setting due to the dominance of true negatives; PR-AUC provides a more informative view by focusing on the rare but critical positive class. R@0.1FPR measures the recall achievable under a 10\% false positive rate, and FPR@0.9R captures the false alarm rate when 90\% of risky cases are recalled. 

\subsection{Overall Performance~(RQ1)}

To address \textbf{RQ1}, we evaluate our proposed method AC-MIL against various baselines on the May and June datasets. The results are presented in Table~\ref{tab:exp:main}, from which we draw the following observations.

\textbf{AC-MIL achieves new state-of-the-art performance.}
AC-MIL consistently outperforms all baselines across both datasets and all metrics, establishing a new performance benchmark for live streaming risk assessment. Specifically, it improves PR-AUC by up to +4.4\% on the May dataset and +4.0\% on the June dataset compared to the strongest baseline. This substantial gain demonstrates the effectiveness of our model in distinguishing risky rooms. Importantly, AC-MIL also achieves significantly lower FPR@0.9R, especially on the May dataset, indicating a stronger ability to reduce false positives while maintaining high recall.

\textbf{MIL-based methods outperform traditional sequence models.}
Among all baselines, we observe that MIL-based methods generally outperform sequence models. This confirms that modeling live streaming rooms as bags of interactions with room-level supervision is more appropriate for risk assessment than simply processing them as flat sequences. Sequence models tend to blur localized high-risk signals when faced with long or noisy interaction histories, leading to degraded performance.
Despite their limitations, the decent performance of sequence models highlights the value of flattened action representations, which motivates our design of a full-sequence Action Field Encoder as the initial step in AC-MIL.

\textbf{AC-MIL further advances MIL approach with structured capsule modeling.}
While the best-performing baseline~(TimeMIL) leverages temporal grouping, it overlooks the underlying spatio-temporal dependencies among participants. AC-MIL addresses this limitation by constructing user–timeslot capsules, which aggregate interactions within localized spatio-temporal contexts. This structured abstraction helps preserve fine-grained risk signals that may be subtle or coordinated. Furthermore, the cross-granular fusion module enables the model to capture collaborative or anomalous behavior patterns across different semantic levels, thereby enhancing its detection capability.



\subsection{Ablation Study~(RQ2)}

To address \textbf{RQ2}, we design four ablated variants of AC-MIL to assess the contribution of individual design branches and the effectiveness of the overall architecture:

\begin{itemize}[leftmargin=*,topsep=5pt]
\item
\textbf{AC-MIL w/o a} removes the Action Field Encoder and directly constructs capsules from raw action embeddings. It also omits the action-level representation in the final fusion.
\item
\textbf{AC-MIL w/o c} replaces the graph-aware self-attention with standard self-attention and excludes the capsule-level room representation during gating.
\item
\textbf{AC-MIL w/o u} removes the user-level modeling branch along with the user-level room representation.
\item
\textbf{AC-MIL w/o t} removes the timeslot-level modeling branch along with the timeslot-level room representation.
\end{itemize}

\begin{table}[b]
\centering
\caption{Ablation study of AC-MIL on June dataset.}
\label{tab:exp:ablation}
\resizebox{\columnwidth}{!}{
  \begin{tabular}{@{}c|cccc@{}} 
  \toprule
  \textbf{} & \textbf{PR-AUC $\uparrow$} & \textbf{F1-score $\uparrow$} & \textbf{R@0.1FPR $\uparrow$} & \textbf{FPR@0.9R $\downarrow$} \\ 
  \midrule
  AC-MIL & \textbf{0.7311} & \textbf{0.6777} & \textbf{0.8546} & \textbf{0.1345} \\
  \midrule
  AC-MIL w/o a & 0.7146 & 0.6569 & 0.8445 & 0.1548 \\
  AC-MIL w/o c & 0.7188 & 0.6617 & 0.8455 & 0.1459 \\
  AC-MIL w/o u& 0.7267 & 0.6559 & 0.8486 & 0.1446 \\
  AC-MIL w/o t & 0.7163 & 0.6733 & 0.8506 & 0.1449 \\
  \bottomrule
  \end{tabular}
}
\end{table}
As shown in Table~\ref{tab:exp:ablation}, removing any core component consistently degrades performance. Specifically, excluding the Action Field Encoder (\textbf{AC-MIL w/o a}) reduces PR-AUC by 2.3\%, underscoring the importance of global action context. Removing relational capsule learning (\textbf{AC-MIL w/o c}) results in a 1.7\% PR-AUC drop, highlighting the effectiveness of our dynamic relation-aware graph in capturing meaningful correlations. Omitting user- or timeslot-level modeling (\textbf{AC-MIL w/o u} and \textbf{AC-MIL w/o t}) also harms performance across all metrics, confirming the complementary nature of the two views. These results validate that AC-MIL’s hierarchical serial-parallel design and multi-view fusion are essential for risk assessment at the live streaming room level.

\subsection{Case Study~(RQ3)}
\begin{figure}[tb]
  \centering
\includegraphics[width=0.48\textwidth]{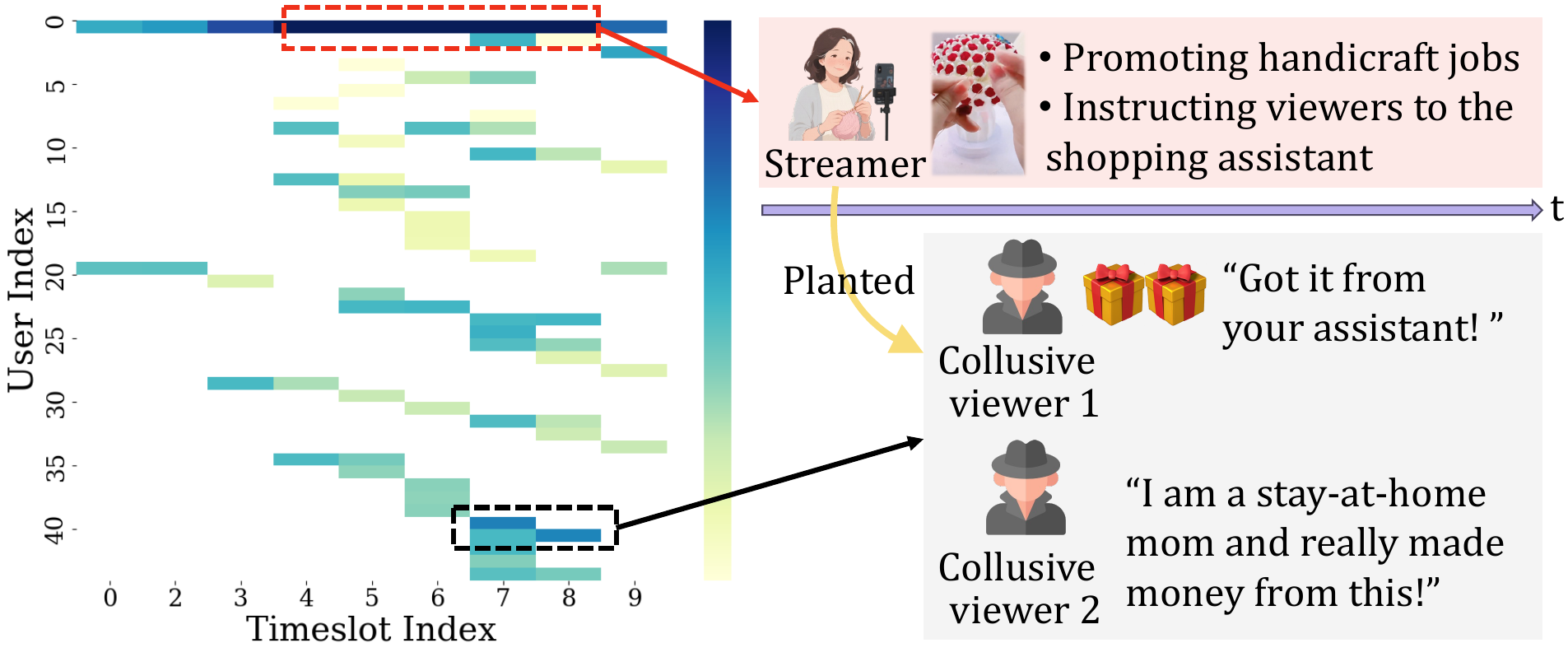}
\caption{
An illustrative case of collusive fraud detected by AC-MIL. 
\textbf{Left:} Attention heatmap over the User-Timeslot capsule space. 
\textbf{Right:} The streamer promotes a fake part-time job, followed by planted viewers. 
Victims are later charged for training and materials before the scammers vanish.
}
    \vspace{-1em}
  \label{fig:exp:case}
\end{figure}

To address \textbf{RQ3}, we present a case study demonstrating how AC-MIL uncovers collusive fraud schemes in live streaming. With attention scores computed as in Eq.~\ref{eq:att score}, as shown in Figure~\ref{fig:exp:case}, AC-MIL assigns high attention to viewer clusters whose activity is temporally synchronized with the streamer’s suspicious promotional behavior, suggesting potential collusion.
More case analyses are in Appendix~\ref{sec:app:case}.

In this example, the streamer actively promotes online handicraft jobs and instructs viewers to consult a shopping assistant for further guidance. Soon after, several planted viewers emerge, sending virtual gifts to attract attention and posting persuasive comments such as “Got it from your assistant!” or “I’m a stay-at-home mom and really made money from this!”, simulating authentic user experiences. These messages are strategically timed to build trust and trigger engagement from real viewers.

By highlighting such coordinated viewer–timeslot patterns, AC-MIL provides actionable cues for detecting suspicious segments and enables timely, evidence-based risk intervention.

\begin{figure}[b]
  \centering
\includegraphics[width=0.45\textwidth]{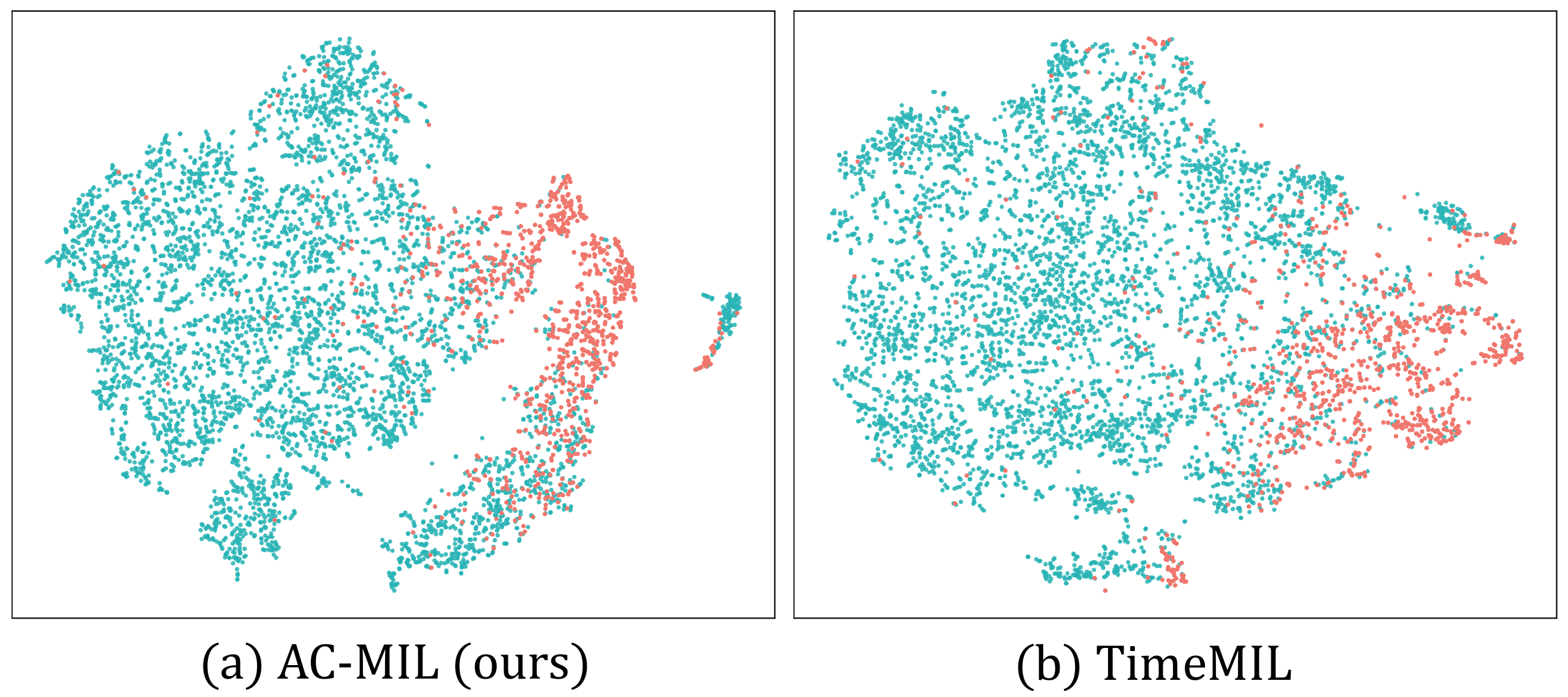}
  \caption{Visualization of room representations learned by AC-MIL and TimeMIL, where salmon nodes represent risky rooms and turquoise nodes represent benign ones.}
    \vspace{-1em}
  \label{fig:exp:vis}
\end{figure}
\subsection{Visualization~(RQ4)}
To address \textbf{RQ4}, we visualize the learned room representations to illustrate how AC-MIL captures discriminative patterns for risk identification.
We apply t-SNE to reduce the dimensionality of room representations learned by AC-MIL and TimeMIL~(the strongest baseline) into two dimensions for visualization. 

As depicted in Figure~\ref{fig:exp:vis}, the representations learned by AC-MIL exhibit a clear clustering structure, where risky and non-risky rooms are well separated.
In contrast, the representations produced by TimeMIL show more overlap between the two classes, indicating less discriminative power.
This clearer separation also aligns with AC-MIL's higher recall and lower false alarm rate, indicating that it effectively distinguishes between risky and benign room patterns.
\subsection{Online Test~(RQ5)}
To address \textbf{RQ5}, we evaluate the performance of AC-MIL against two deployed online models on more up-to-date data. The XGBoost-based model relies on hand-crafted features derived from statistics of streamers’ historical activities and real-time room indicators to make a preliminary judgment.
And the Transformer-based model flattens the entire room-level action sequence as input, but does not explicitly model user–timeslot granularity. Table~\ref{tab:exp:online} shows that AC-MIL consistently outperforms both online models across all metrics, demonstrating superior performance. 

In particular, besides significant increases in PR-AUC, AC-MIL also achieves higher constrained recall (R@0.1FPR), outperforming XGBoost by +44.46\% and Transformer by +6.7\%. Moreover, AC-MIL reduces FPR@0.9R to 0.1655, significantly lower than XGBoost and Transformer, which is crucial for minimizing false alarms in online settings for preserving normal users' experience.

\begin{table}[t]
\centering
\caption{Online test between AC-MIL and deployed models (XGBoost-based and Transformer-based) on real-world data spanning from 07/13/25 to 07/14/25. A 1:10 positive-to-negative sampling ratio is applied.}

\label{tab:exp:online}
\resizebox{\columnwidth}{!}{
  \begin{tabular}{@{}c|cccc@{}} 
  \toprule
  \textbf{} & \textbf{PR-AUC $\uparrow$} & \textbf{F1-score $\uparrow$} & \textbf{R@0.1FPR $\uparrow$} & \textbf{FPR@0.9R $\downarrow$} \\ 
  \midrule
  AC-MIL & \textbf{0.7355} & \textbf{0.6574} & \textbf{0.8390} & \textbf{0.1655} \\
  \midrule
 XGBoost & 0.4512 & 0.4479 & 0.5808 & 0.5514 \\
 Transformer & 0.6381 & 0.6097 & 0.7864 & 0.1909 \\
  \bottomrule
  \end{tabular}
}
\end{table}

\subsection{Hyperparameter Sensitivity Test~(RQ6)}
To address \textbf{RQ6}, we conduct a sensitivity analysis on the June dataset to examine how AC-MIL responds to key hyperparameters used in capsule construction,
including the timeslot length and the number of selected top viewers.

As shown in Table~\ref{tab:exp:hyper}, AC-MIL demonstrates stable performance
across a range of hyperparameter settings. Varying the timeslot length from
50s to 150s leads to only marginal changes in evaluation metrics, indicating
that the model is not overly sensitive to temporal granularity. Similarly,
adjusting the number of top viewers from 30 to 70 results in minor performance
fluctuations, with no configuration causing substantial performance degradation.

Overall, these results suggest that AC-MIL is robust to reasonable variations
in capsule construction hyperparameters.

\begin{table}[htbp]
\centering
\small
\caption{Sensitivity analysis of AC-MIL to hyperparameters
(timeslot length and number of selected top viewers) on the June dataset.
The (100s, 50 viewers) setting is the default configuration used in all main experiments.}
\label{tab:exp:hyper}
\resizebox{\columnwidth}{!}{
\begin{tabular}{c| c c c c c}
\toprule
\textbf{Timeslot~(s)} & \textbf{100} & 100 & 100 & 50 & 150 \\
\textbf{\#Top Viewers} & \textbf{50} & 30 & 70 & 50 & 30 \\
\midrule
\textbf{PR-AUC $\uparrow$} & \textbf{0.7311} & 0.7157 & 0.7279 & 0.7308 & 0.7312 \\
\textbf{F1-score $\uparrow$} & \textbf{0.6777} & 0.6784 & 0.6657 & 0.6739 & 0.6781 \\
\textbf{R@0.1FPR $\uparrow$} & \textbf{0.8546} & 0.8646 & 0.8596 & 0.8636 & 0.8497 \\
\textbf{FPR@0.9R $\downarrow$} & \textbf{0.1345} & 0.1270 & 0.1250 & 0.1357 & 0.1375 \\
\bottomrule
\end{tabular}
}
\end{table}

\section{Conclusion}\label{sec:conclude}
In this work, we presented a pioneering study on risk assessment in live streaming rooms under weak supervision, where only room-level labels were available. We formulated the problem as an MIL task by defining \emph{user–timeslot capsules} that capture localized user behaviors within temporal windows. 
To effectively model the complex temporal and cross-user dependencies inherent in live streaming risks, we proposed AC-MIL, an Action-Aware Capsule MIL framework. It encodes user actions into semantic capsules, then models their relational dependencies through adaptive reasoning, and aggregates multi-granular contextual information to deliver accurate and interpretable risk predictions.

Extensive experiments on large-scale industrial datasets from Douyin demonstrated that AC-MIL set new SOTA performance by a large margin, significantly outperforming existing MIL and sequential baselines with notably high recall and low false alarm rates for early-stage risk detection. 
Notably, our method provides localized capsule-level interpretability, enabling the identification of problematic segments as actionable evidence to support targeted risk interventions.

\begin{acks}
The research work is supported by the National Natural Science Foundation of China under Grant Nos. U2436209, 62576333, and 62406307, the Strategic Priority Research Program of the Chinese Academy of Sciences under Grant No. XDB0680201, the Beijing Natural Science Foundation under Grant No. JQ25015, and the Innovation Funding of ICT, CAS under Grant No. E461060.
\end{acks}

\bibliographystyle{ACM-Reference-Format}
\bibliography{refs}

\appendix
\begin{figure*}[t]
  \centering
  \includegraphics[width=0.9\textwidth]{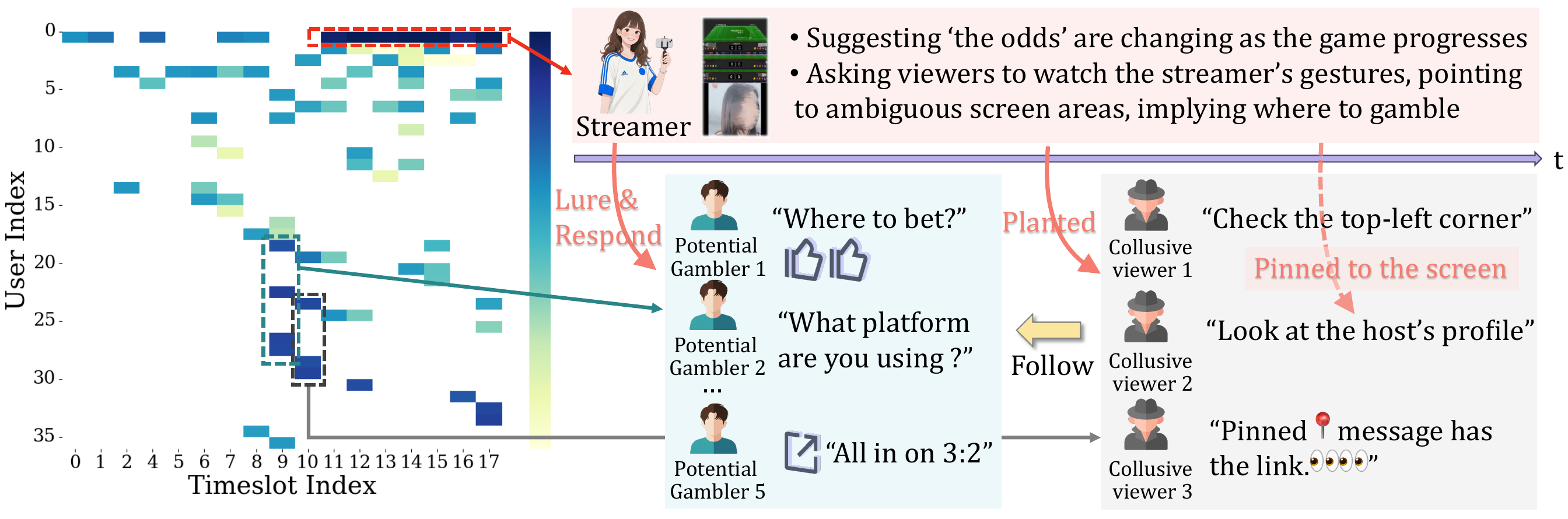}

  \caption{An illustrative case of covert gambling promotion detected by AC-MIL. Left: Attention heatmap over the User-Timeslot capsule space. Right: High-attention regions reveal a set of risky capsules: The streamer leverages ambiguous gestures and subtle hints to attract potential gamblers, who begin to inquire about betting. Later, collusive viewers indirectly suggest where to find the gambling link (also acknowledged by the streamer), forming a case of off-platform gambling redirection.
}
  \label{fig:app:case:bet}
\end{figure*}
\section{Experimental Setup Details}
\subsection{Data Collection and Preprocessing}
\label{sec:app:data}

We collect various actions from viewers and streamers.
The viewer action space encompasses various types of interactions, including room entry, comments, chat highlights, leaderboard appearances, danmaku messages, gifting, likes, shares, co-stream requests, and group joins. 
For streamer-side activities, we consider the start of the stream, voice-to-text transcribed speech, and OCR-based visual content monitoring. The latter two are obtained through scheduled platform inspections, where the raw data is discretized at collection time as part of the risk control sampling process. The action sequences are capped at 2,096 tokens in length. 

Textual fields within actions (e.g., comment content or spoken scripts) are pre-encoded using a Chinese-BERT\footnote{https://huggingface.co/google-bert/bert-base-chinese} encoder. All interaction data mentioned above is collected from publicly available content and processed in compliance with platform privacy and security policies.
\subsection{Baselines}
\label{sec:app:baselines}
Our baseline methods fall into two categories:
(i)~\textit{Sequence Models} to directly model the action sequences of rooms: 
\begin{itemize}[leftmargin=*,topsep=5pt]
\item
\textbf{Transformer}~\cite{vaswani2017attention} is a widely used sequence model with a self-attention mechanism. 
\item 
\textbf{Reformer}~\cite{kitaev2020reformer} is a Transformer variant with efficient locality-sensitive hashing. 
\item 
\textbf{Informer}~\cite{zhou2021informer} is another efficient Transformer variant employing sparse attention and distillation for scalable long-sequence modeling.
\end{itemize}
(ii) \textit{MIL methods} to learn room-level predictions via instance aggregation: 
\begin{itemize}[leftmargin=*,topsep=5pt]
\item
\textbf{mi-NET}~\cite{wang2018revisiting} directly aggregates instance scores into the final bag prediction.
\item
\textbf{Attention MIL~(AtMIL)}~\cite{ilse2018attention} learns soft attention weights to highlight informative instances. 
\item 
\textbf{Additive MIL~(AdMIL)}~\cite{javed2022additive} proposes additive pooling, which combines attention and instance pooling in an additive manner to capture both importance and presence of instances.
\item
\textbf{MIL-LET}~\cite{early2024inherently} decouples attention and classification by computing them independently and combining them via attention-weighted prediction scores. 
\item 
\textbf{TimeMIL}~\cite{chen2024timemil} is a time-aware MIL pooling method with learnable wavelet positional tokens. 
\item 
\textbf{TAIL-MIL}~\cite{jang2025tail} applies 2D-MIL to multivariate time series.
\end{itemize}
\begin{figure*}[t]
\centering
\includegraphics[width=0.9\textwidth]{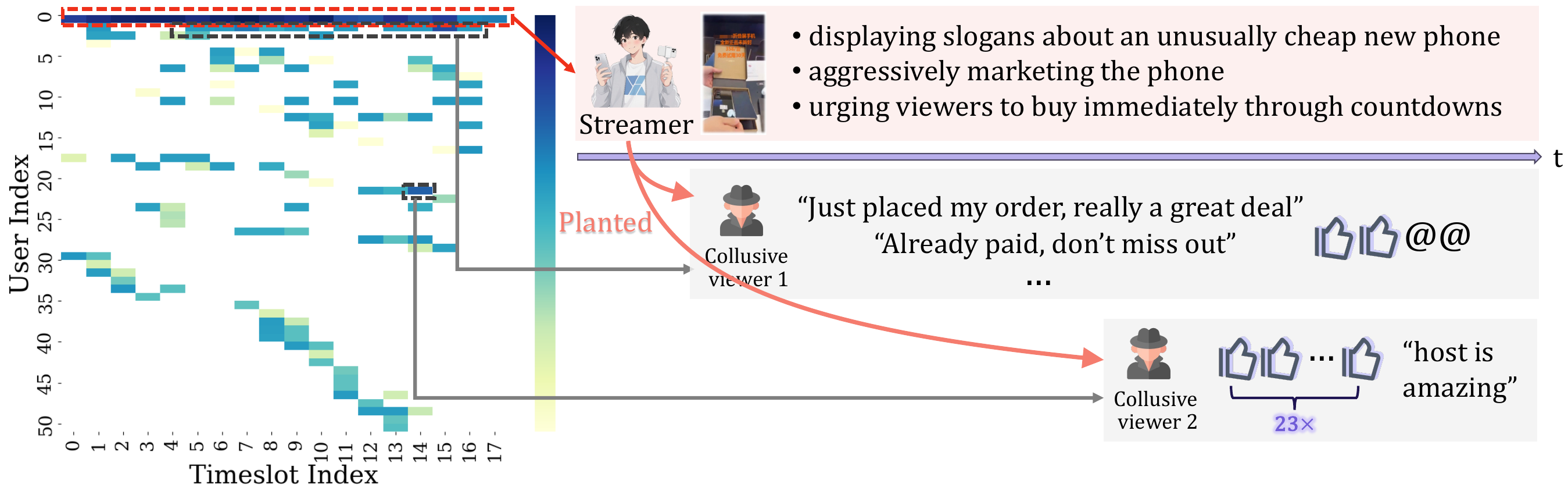}
\caption{An illustrative case of fraudulent cheap phone sales detected by AC-MIL. \textbf{Left}: Attention heatmap over the User–Timeslot capsule space. \textbf{Right}: Risky capsules reveal a coordinated scheme centered on an unrealistically low-priced phone scam. The streamer aggressively promotes the phone with urgency tactics, while fake “buyers” generate bursts of likes and positive comments within short time frames, creating false trust and pushing real users toward payment fraud.}
\label{fig:app:case:phone}
\end{figure*}

\section{Further Case Studies}
\label{sec:app:case}
In the following appendix, we provide more case studies that further demonstrate the value of capsule-level interpretability. Our model not only distinguishes important capsules across users and time but also captures meaningful sequences of actions within each capsule. This two-level insight helps reveal coordinated and evolving fraudulent behaviors more clearly, which is useful for practical risk detection and intervention.

\noindent
\textbf{Illegal gambling promotion.}
Here, we present a case of illegal gambling promotion in live streaming. As illustrated in Figure~\ref{fig:app:case:bet}, AC-MIL highlights some suspicious promotional segments in a risky streaming room centered around gambling-related activities. In this session, the streamer continuously displayed live scores from multiple soccer matches, rapidly switching between games without offering any commentary, thereby cultivating an atmosphere of intense betting interest.

Although the streamer never explicitly mentioned gambling, subtle cues were embedded in the stream, such as saying “odds are changing fast” and repeated gestures pointing toward ambiguous screen areas. These vague prompts appeared sufficient to trigger some viewers to ask questions such as “Where to bet?” and “Which platform are you using?”, and a user even forwarded this live room and specified the exact betting odds like “All in on 3:2”.

After these viewer inquiries, several accounts responded with suggestive and directional messages such as “Check the top-left corner,” “Look at the host’s profile,” or “The pinned message has the link.” The speed and consistency of these replies strongly suggest that these accounts were pre-arranged, deliberately planted to steer genuine users toward off-platform gambling.

In contrast to the prior case involving bot-like coordinated fraud, most suspicious participants here appeared to be real users who were gradually lured into risky behavior. Their progression was facilitated by a small group of strategically placed collusive audience members. This case illustrates how risk can emerge organically from subtle yet persistent cues, then escalate rapidly through collusive interactions, making it particularly challenging to detect under weak supervision.

AC-MIL effectively captures this evolving interaction pattern, assigning high attention to both the streamer’s ambiguous prompts and the subsequent user responses. It surfaces interpretable signals of escalating intent and coordination, providing actionable evidence to support timely intervention.

\noindent
\textbf{Fraudulent phone sales.}
Then, we present a case of fraudulent phone sales in live streaming. As shown in Figure~\ref{fig:app:case:phone}, AC-MIL flags a suspicious room where the screen prominently displays slogans like “Brand-new folding-screen smartphone, only 358 RMB,” while the streamer aggressively markets it as a “flagship configuration” and claims “factory clearance—only a few left.” The broadcast is accompanied by countdown timers and constant verbal cues urging immediate purchase.

The streamer’s on-screen promotions and verbal cues are clearly malicious, and coordinated actions by some viewers make the case even more suspicious. For example, one viewer often posts buyer-like comments such as “Just placed my order, really a great deal” and “Already paid, don’t miss out,” while frequently liking the stream and tagging friends. Another viewer repeatedly likes the stream over 20 times in a short period and posts simple praise like “host is amazing,” which is unusual and clearly meant to boost popularity.

These actions form clear patterns that our capsule-based method captures. One capsule may include more than 20 likes from the same user within a short time, which signals suspicious activity. By grouping related actions into capsules, the model detects coordinated or unnatural behavior. Unlike traditional methods that analyze actions separately or rely on static user data, AC-MIL uses capsules as the basic units for detection, allowing it to catch subtle and timely coordination.

In this case, AC-MIL assigns high attention to the streamer’s capsule and two viewer capsules, highlighting their coordinated role in the scam. This shows how capsule-level modeling uncovers meaningful behavior patterns behind unusually low-priced promotions.

\end{document}